\documentclass[lettersize,journal]{IEEEtran}
\usepackage{amsmath,amsfonts}
\usepackage{algorithmic}
\usepackage{algorithm}
\usepackage{array}
\usepackage[caption=false,font=normalsize,labelfont=sf,textfont=sf]{subfig}
\usepackage{textcomp}
\usepackage{stfloats}
\usepackage{url}
\usepackage{verbatim}
\usepackage{graphicx}
\usepackage{cite}

\usepackage{hyperref}
\usepackage{xcolor}
\usepackage{booktabs}
\usepackage{multirow}
\definecolor{highlightgray}{gray}{0.9}
\usepackage{colortbl}

\usepackage{xcolor}   
\usepackage{amssymb} 
\usepackage{pifont}

\begin{document}

\title{Reasoning-Driven Amodal Completion: Collaborative Agents and Perceptual Evaluation}

\author{Hongxing Fan, Shuyu Zhao, Jiayang Ao and Lu Sheng, \textit{Member, IEEE}
\thanks{Manuscript received 23 December, 2025; This work was supported by National Natural Science Foundation of China (62132001), and the Fundamental Research Funds for the Central Universities. \textit{(Corresponding author: Lu Sheng.)}

Hongxing Fan is with the School of Computer Science and Engineering, Beihang University, Beijing 100191, China (e-mail: fanhongxing@buaa.edu.cn).

Shuyu Zhao and Lu Sheng are with the School of Software, Beihang University, Beijing 100191, China (e-mail: zhaoshuyu@buaa.edu.cn, 
lsheng@buaa.edu.cn).

Jiayang Ao is with the School of Computing and Information Systems, The University of Melbourne, Victoria 3053, Australia (e-mail: jiayanga@student.unimelb.edu.au).}}


\markboth{\tiny{This work has been submitted to the IEEE for possible publication. Copyright may be transferred without notice, after which this version may no longer be accessible.}}%
{Shell \MakeLowercase{\textit{et al.}}: A Sample Article Using IEEEtran.cls for IEEE Journals}


\maketitle
\begin{abstract}
Amodal completion, the task of inferring invisible object parts, faces significant challenges in maintaining semantic consistency and structural integrity. Prior progressive approaches are inherently limited by \textit{inference instability} and \textit{error accumulation}. To tackle these limitations, we present a Collaborative Multi-Agent Reasoning Framework that explicitly decouples Semantic Planning from Visual Synthesis. By employing specialized agents for upfront reasoning, our method generates a structured, explicit plan before pixel generation, enabling visually and semantically coherent single-pass synthesis. We integrate this framework with two critical mechanisms: (1) a self-correcting Verification Agent that employs Chain-of-Thought reasoning to rectify visible region segmentation and identify residual occluders strictly within the Semantic Planning phase, and (2) a Diverse Hypothesis Generator that addresses the ambiguity of invisible regions by offering diverse, plausible semantic interpretations, surpassing the limited pixel-level variations of standard random seed sampling. Furthermore, addressing the limitations of traditional metrics in assessing inferred invisible content, we introduce the MAC-Score (MLLM Amodal Completion Score), a novel human-aligned evaluation metric. Validated against human judgment and ground truth, these metrics establish a robust standard for assessing structural completeness and semantic consistency with visible context. Extensive experiments demonstrate that our framework significantly outperforms state-of-the-art methods across multiple datasets. Our project is available at: https://fanhongxing.github.io/remac-page.
\end{abstract}

\section{Introduction}
\begin{figure*}
    \centering
    \includegraphics[width=\textwidth]{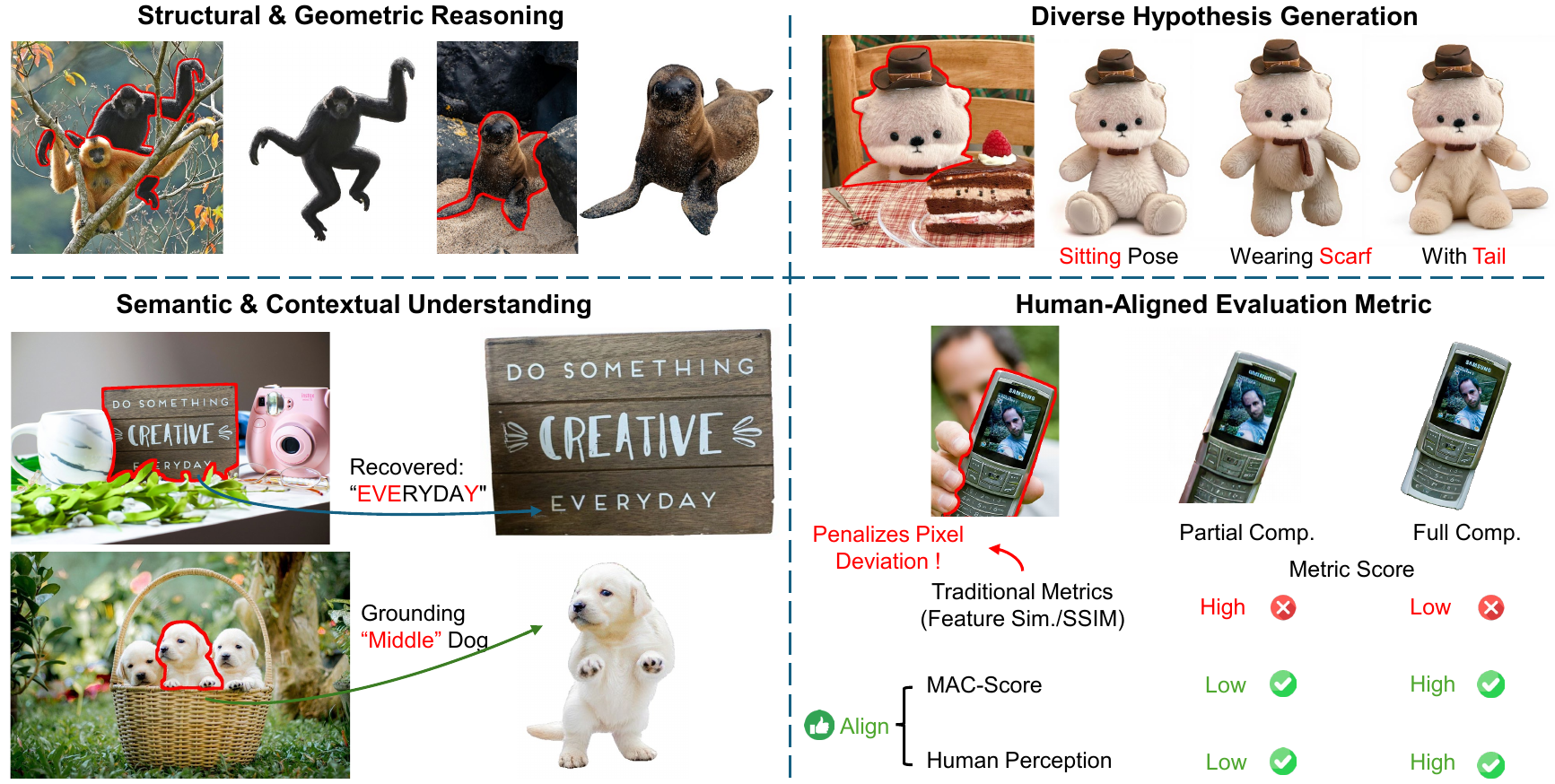}
    \caption{
    Our framework tackles complex occlusions through these key capabilities: 
    (1) \textbf{Structural \& Semantic Reasoning}, which recovers geometric continuity (e.g., hidden limbs) and contextual details (e.g., text) beyond pixel clues; and 
    (2) \textbf{Diverse Hypothesis Generation}, which models the multimodal nature of invisible regions (e.g., diverse plushie states). 
    Furthermore, we introduce
    (3) the \textbf{MAC-Score}, a human-aligned evaluation metric. 
    As shown in the bottom-right, it resolves the mismatch where incomplete results are favored by traditional metrics, providing a robust standard for amodal completion.
    }
    \label{fig:teaser}
\end{figure*}

\IEEEPARstart{A}{modal} completion, the ability to perceive and reconstruct the invisible parts of partially occluded objects, stands as a cornerstone of visual intelligence~\cite{kanizsa1979organization, ao2023image}. 
While humans effortlessly infer global structure from partial observations, replicating this cognitive capability in computational systems remains a formidable challenge. 
Advancements in this domain empower intelligent image editing, augmented reality, and digital asset generation\cite{xu2024amodal}. 
Crucially, the problem extends beyond mere pixel inpainting; it requires a sophisticated synthesis of global geometric reasoning, semantic context interpretation, and the plausible inference of unseen content.

\begin{figure}[t]
    \centering
    \includegraphics[width=1.0\linewidth]{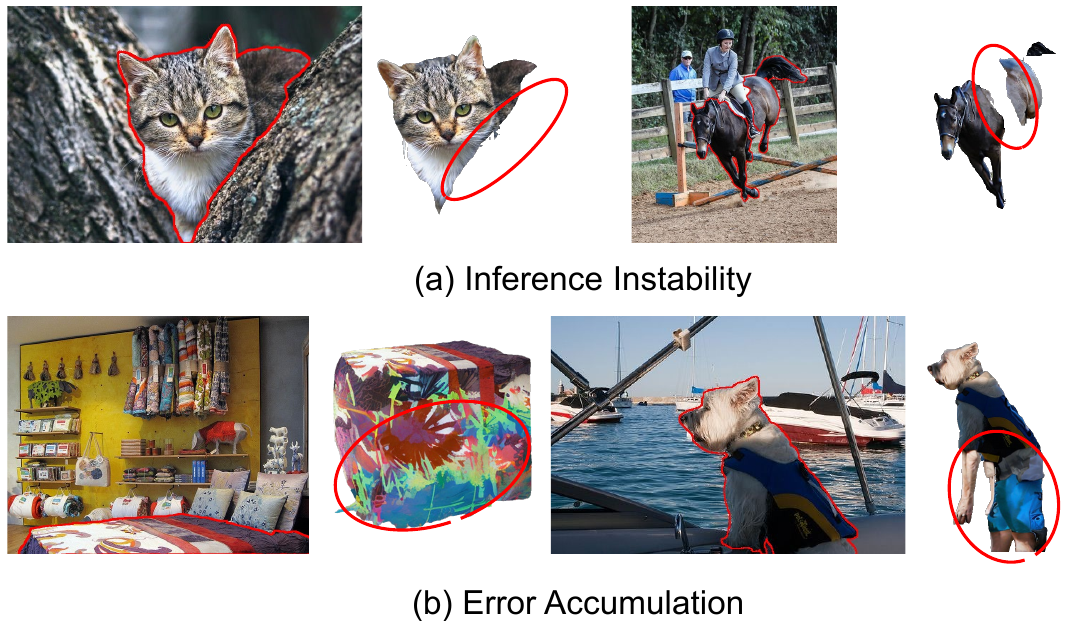}
    \caption{\textbf{Common failure modes of progressive methods.} 
    (a) \textbf{Inference Instability}: The progressive process often terminates prematurely due to a lack of global planning, resulting in incomplete or truncated objects.
    (b) \textbf{Error Accumulation}: Early-stage errors propagate and amplify through iterative steps, causing structural inconsistencies and artifacts. }
    \label{fig:failure_modes}
\end{figure}

Existing approaches~\cite{ehsani2018segan, yan2019visualizing, ling2020variational, zhou2021human, bowen2021oconet, ozguroglu2024pix2gestalt, zhan2020self, reddy2022walt, liu2024object} in this domain predominantly rely on training-based pipelines. While effective in constrained settings, these methods depend heavily on large, task-specific datasets and often fail to generalize to complex occlusion scenarios encountered in real-world. More recent training-free methods~\cite{xu2024amodal, ao2025open} seek to address these limitations by leveraging pre-trained generative priors. However, most of these approaches adopt a progressive iterative strategy that is vulnerable to two critical failure modes: \textit{inference instability} and \textit{error accumulation}. The former refers to the tendency of the iterative process to terminate prematurely or collapse structurally, yielding fragmented results. The latter occurs when minor early-stage artifacts or semantic drifts propagate and amplify through subsequent steps, progressively degrades compromising global consistency (see Figure~\ref{fig:failure_modes}).

\begin{figure}[t]
    \centering
    \includegraphics[width=1.0\linewidth]{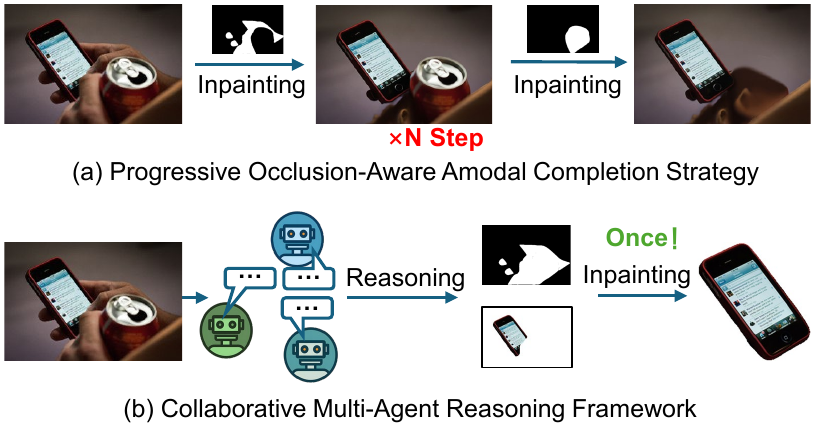}
    \caption{\textbf{Comparison of amodal completion paradigms.} (a) \textbf{Progressive Strategy}: Iterative expansion ($\times N$) is vulnerable to error accumulation. (b) \textbf{Our Framework}: A holistic ``reason-then-synthesize'' approach. By determining the comprehensive plan upfront, we achieve robust single-pass synthesis, ensuring global consistency without iterative instability.}
    \label{fig:comparison_paradigm}
\end{figure}

To address these challenges, our preliminary work~\cite{fan2025multi} introduced a novel \textbf{Collaborative Multi-Agent Reasoning Framework}. Departing from the traditional pixel-level iterative paradigm, we reformulate amodal completion as a cognitive reasoning task that necessitates a clear separation between ``thinking'' and ``drawing'' (see Figure~\ref{fig:comparison_paradigm}). Specifically, our framework explicitly decouples the process into two phases: \textbf{Semantic Planning} and \textbf{Visual Synthesis}. 
During \textbf{Semantic Planning}, multiple specialized Multimodal Large Language Model (MLLM) agents collaboratively analyze the scene by disentangling occlusion relationships, determining necessary boundary expansion, and inferring semantic attributes from the global context.
This yields a comprehensive execution plan, comprising an inpainting mask and a fine-grained description, which is finalized prior to the visual synthesis phase. Subsequently, the \textbf{Visual Synthesis} phase leverages this plan to generate the final photorealistic result. By shifting to this holistic paradigm, our method circumvents the \textit{inference instability} and \textit{error accumulation} in progressive execution, achieving state-of-the-art (SOTA) performance.

While our preliminary study effectively mitigates \textit{inference instability} and \textit{error accumulation}, it leaves critical challenges in reasoning robustness, semantic ambiguity, and evaluation validity unaddressed. In this work, we systematically advance the pipeline into a robust, closed-loop, and ambiguity-aware framework to tackle these challenges. 
First, to address the potential perceptual oversights in the initial analysis, we introduce a Chain-of-Thought (CoT)~\cite{wei2022chain} \textbf{Verification Agent}. This self-correcting mechanism employs CoT reasoning to correct visible region inaccuracies and identify residual occluders strictly within the planning phase. Second, we distinguish between semantic ambiguity and low-level stochasticity. Since existing random-seed approaches fail to capture the meaningful diversity of the ill-posed solution space, we propose a \textbf{Hypothesis Generator}. Leveraging MLLMs, this module produces diverse, interpretable semantic hypotheses rather than a single deterministic outcome. Finally, we resolve the evaluation mismatch where traditional metrics (e.g., LPIPS) can unintentionally ``reward incompleteness.'' We establish the \textbf{MAC-Score} (MLLM Amodal Completion Score), a human-aligned metric that shifts assessment from rigid pixel matching to structural completeness and semantic consistency.

In summary, our main contributions are:

\begin{itemize}
\item We propose a robust \textbf{Closed-Loop Collaborative Multi-Agent Reasoning Framework} that decouples semantic planning from visual synthesis, while integrating a CoT Verification mechanism for reliable error handling. This synergistic design enabling structural completeness and semantic consistency, yielding SOTA performance.

\item We propose \textbf{Diverse Hypothesis Generation} to explicitly handle the inherent ambiguity of invisible regions. 
Our framework enumerates and ranks multiple interpretable semantic hypotheses for the occluded content, enabling meaningful and controllable diversity beyond pixel-level variations induced by random seeds.

\item We establish the \textbf{MAC-Score}, a robust perceptual evaluation paradigm to address the limitations of traditional metrics which often penalize plausible completions. This new standard, comprising the MAC-Completeness and MAC-Consistency, aligns strongly with human intuition and offers a reproducible benchmark for the community.

\end{itemize}

\section{Related Work}
\textbf{Amodal Completion} addresses the reconstruction of invisible parts in occluded objects~\cite{kanizsa1979organization}. Existing approaches fall into two main paradigms: training-based and training-free.
Training-based methods typically employ pipelines that predict occlusion masks prior to inpainting~\cite{ehsani2018segan, yan2019visualizing, ling2020variational, zhou2021human, bowen2021oconet, ozguroglu2024pix2gestalt} or exploit self-supervised structural priors~\cite{zhan2020self, reddy2022walt, liu2024object}. 
Recent works in computer graphics have extended these paradigms to 3D data~\cite{zhou2025amodalgen3d,wu2025unsupervised,wu20243d}, employing multi-view adversarial learning~\cite{wu2025unsupervised} or prior-assisted weak supervision~\cite{wu20243d} to complete point clouds, specifically targeting the challenge of unseen categories.
However, these methods heavily rely on task-specific datasets. Due to the scarcity of real-world amodal annotations, training often resorts to synthetic compositing (e.g., pasting patches to simulate occlusion)~\cite{zhou2021human, yan2019visualizing, ozguroglu2024pix2gestalt}. This reliance restricts generalization and robustness in complex, out-of-distribution scenarios, where occlusion patterns, object semantics, and contextual cues differ substantially from synthetic compositions. A limitation persisting even in advanced diffusion-based variants~\cite{liu2024object, ozguroglu2024pix2gestalt}.
Conversely, training-free methods leverage the priors of large pre-trained generative models to bypass specific training~\cite{xu2024amodal, ao2025open}. 
These approaches predominantly adopt progressive strategies, iteratively expanding content within occluded regions. While effective in leveraging open-world knowledge, this step-by-step nature is inherently vulnerable to \textit{inference instability} and \textit{error accumulation}. Alternatively, one-step wide-masking often induces semantic drift or hallucinates occluders instead of targets.
To address these limitations, our work introduces a collaborative multi-agent framework that decouples reasoning from synthesis to achieve reliable, semantically coherent completion.

\noindent\textbf{Multi-Agent Systems for Visual Reasoning and Synthesis.} 
Multi-Agent Systems (MAS) tackle complex vision tasks via a ``divide and conquer'' strategy, typically utilizing a centralized planner to coordinate specialized agents~\cite{tran2025multi,li2025mccd,zhang2025pixelcraft}. 
In visual synthesis, most systems adopt iterative strategies~\cite{mao2025emoagent,ma2025talk2image}. However, this step-by-step paradigm suffers from inherent error accumulation, where early-stage artifacts propagate to compromise the final output~\cite{fan2025multi}.
To mitigate this, our prior work~\cite{fan2025multi} decoupled semantic planning from visual synthesis, employing upfront collaborative planning to enable robust single-pass generation. 
While this effectively eliminates synthesis-stage error propagation, it relies on an open-loop planning phase that lacks mechanisms for self-correction and ambiguity handling. 
This work advances that foundation by introducing closed-loop verification and diverse hypothesis generation, elevating the framework from an efficient execution pipeline to a robust, intelligent reasoning system.

{\bf{Evaluation of Amodal Completion}}
Evaluating amodal completion is challenging due to the lack of ground truth in real images\cite{ao2023image,ao2025open}. While mIoU\cite{qi2019amodal,ling2020variational,ozguroglu2024pix2gestalt} can assess geometric accuracy when masks are available, general evaluation relies on proxy metrics like LPIPS\cite{zhang2018unreasonable}, SSIM\cite{wang2004image}, VGG perceptual distance\cite{gatys2016image}, FID\cite{heusel2017gans}, and CLIP Score\cite{radford2021learning}. However, these proxies fail to directly assess the generated invisible content. They  often check only the preservation of visible parts, which can misleadingly reward failed completions, or rely on flawed assumptions about texture homogeneity\cite{ding2020image}. Even holistic scores like FID and CLIP cannot validate if the generated content is contextually consistent with the visible parts\cite{kim2025preserve}. This gap necessitates costly and subjective human studies, highlighting a critical need for robust automated metrics\cite{ao2025open,fan2025multi}. Our work addresses this gap by proposing a new evaluation paradigm that leverages MLLMs to automate structural completeness and semantic consistency.
\section{Method}

\begin{figure*}[t]
    \centering
    \includegraphics[width=1.0\linewidth]{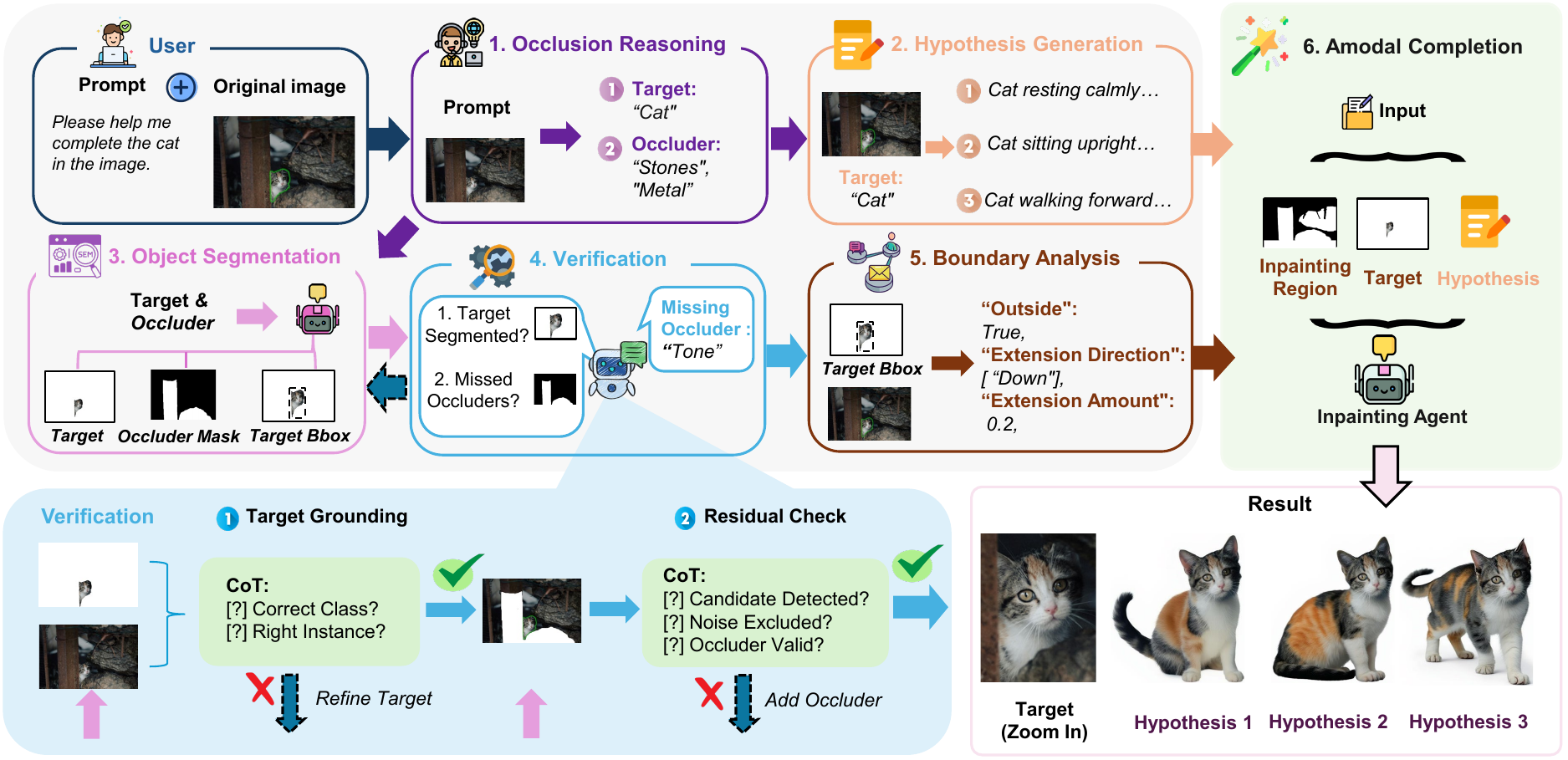}
    \caption{\textbf{Overview of the proposed Closed-Loop Collaborative Multi-Agent Reasoning Framework.} 
    The pipeline decouples semantic planning from visual synthesis through three stages: 
    (1) \textbf{Holistic Collaborative Reasoning}: A coalition of agents synergizes to parse the scene's geometry, forming an initial spatial plan. 
    (2) \textbf{Closed-Loop Verification}: A self-correcting mechanism where a Verification Agent scrutinizes the initial plan to correct segmentation errors and identify residual occluders. 
    (3) \textbf{Hypothesis Generation}: The Hypothesis Agent generates multiple semantic descriptions for the invisible regions to capture diverse plausible interpretations.
    Finally, the \textbf{Inpainting Agent} executes the verified plan to synthesize the high-fidelity amodal result in a single pass.}
    \label{fig:framework_overview}
\end{figure*}

\subsection{Problem Formulation and Overview}
\label{sec:Problem_Formulation}
Amodal completion is the task of recovering the complete geometry and appearance of an object from its partially observed state. Given an observed image $I_{obs} \in \mathbb{R}^{H \times W \times 3}$ containing a target object partially occluded by a set of occluders $\mathcal{O}$ or truncated by the image boundary, the goal is to synthesize a completed image $I_{comp}$ where the invisible regions are plausibly inferred while maintaining consistency with the visible part. 
Unlike standard inpainting tasks supervised by a unique ground truth, amodal completion is inherently ill-posed. To address this ambiguity, we condition the synthesis on the inferred latent semantic variable $S$. The objective is to sample from the posterior distribution $P(I_{comp} | I_{obs}, S)$, enabling the generation of diverse plausible realizations.

We propose a Collaborative Multi-Agent Reasoning Framework. Departing from the iterative generation pipeline, we decouple the process into two distinct phases: \textbf{Semantic Planning} and \textbf{Visual Synthesis}. As illustrated in Figure~\ref{fig:framework_overview}, our framework first employs a coalition of MLLM agents to reason, verify, and formulate a holistic completion plan $\Pi$. This plan comprises a verified inpainting mask $M_{inpaint}$ that delineates the unknown regions, the preserved visible part $I_{vis}$ for structural consistency, and an inferred semantic description $T$. This comprehensive plan is then executed by the Inpainting Agent to produce the final high-fidelity result in a single pass.

\subsection{Holistic Collaborative Reasoning}
\label{sec:holistic_reasoning}

The foundation of our framework is the construction of a robust spatial-semantic plan prior to any pixel synthesis. This stage employs a Coalition of specialized MLLM agents to dissect the scene's geometric context. Let $q$ denote the user query specifying the target object.

\subsubsection{Contextual Analysis and Occlusion Reasoning}
To decompose the scene layout, the \textbf{Occlusion Analysis Agent} ($\mathcal{A}_{occ}$) parses $I_{obs}$ given query $q$. It disentangles depth ordering to identify the set of occluders $\mathcal{O}_{ids}$ obstructing the target:

\begin{equation}
    \mathcal{O}_{ids} = \mathcal{A}_{occ}(I_{obs}, q)
\end{equation}
This structural analysis isolates the target from surrounding occlusions, providing a clean geometric basis for the subsequent segmentation and boundary analysis.

\subsubsection{Geometric Planning and Mask Derivation}
With the occlusion context established, the framework proceeds to ground the identified entities into precise spatial representations. We employ a \textbf{Segmentation Agent} ($\mathcal{A}_{seg}$), utilizing an open-vocabulary segmentation model, to map inferred object identities into the pixel space.
Conditioned on the set of occluders $\mathcal{O}_{ids}$ identified in the previous step, this agent outputs the visible mask of the target, $M_{vis}$, along with the masks for each occluder, $\{M_{occ}^{(i)}\}$.
\begin{equation}
    (M_{vis}, \{M_{occ}^{(i)}\}) = \mathcal{A}_{seg}(I_{obs}, \mathcal{O}_{ids}, q)
\end{equation}

To address potential boundary truncation where the object extends beyond the image, we employ a \textbf{Boundary Analysis Agent} ($\mathcal{A}_{bdy}$) that adopts a hybrid reasoning strategy. By analyzing the geometric alignment of $M_{vis}$ with the image borders alongside the visual context of $I_{obs}$, the agent infers the extent of the invisible truncation. It estimates a parameter vector $\mathbf{e} = [e_t, e_b, e_l, e_r]$, representing the relative expansion ratios for the top, bottom, left, and right margins, respectively:
\begin{equation}
    \mathbf{e} = \mathcal{A}_{bdy}(I_{obs}, M_{vis})
\end{equation}
This step ensures the canvas is dynamically expanded to accommodate the full extrapolated geometry.

\subsubsection{Initial Geometric Planning}
The final step of this stage consolidates the derived spatial components into a preliminary inpainting mask $M_{inpaint}$. To mitigate boundary artifacts at occlusion interfaces, morphological dilation ($\oplus$) with a structuring element $B$ is applied to each occluder mask to ensure the synthesized region slightly overlaps with the occlusion boundary. This is aggregated with the expansion region $M_{exp}$, which corresponds to the canvas extension defined by $\mathbf{e}$:
\begin{equation}
    M_{inpaint} = \left( \bigcup_{i} (M_{occ}^{(i)} \oplus B) \right) \cup M_{exp}
    \label{eq:mask_union}
\end{equation}
The resulting mask $M_{inpaint}$ establishes the initial spatial constraints to be scrutinized by the verification phase.

\subsection{Closed-Loop Verification via Chain-of-Thought}
\label{sec:refinement}

The open-loop reasoning in the initial stage relies on the accuracy of the collaborative analysis. However, perceptual errors may occur, potentially leading to a flawed plan. To ensure reliability, we introduce a closed-loop Verification Mechanism to validate the results before synthesis.

We deploy a \textbf{Verification Agent} ($\mathcal{A}_{ver}$) to scrutinize the preliminary results. First, it performs a \textbf{Target Grounding} to verify if the target object has been successfully segmented. Should this check fail, the Segmentation Agent is immediately triggered to re-segment the target based on the original query.
Subsequently, the verification agent proceeds to the \textbf{Residual Check}. To facilitate this, we employ a \textbf{``White-Out'' Strategy} where all currently identified occluders are masked with pure white in the input image, eliminating visual redundancy. 

Critically, the agent operates under a Programmatic CoT protocol ($\mathcal{P}_{cot}$) that enforces a strict three-step reasoning:
\begin{enumerate}
    \item \textbf{Candidate Identification:} The agent scans the processed image to list all potential objects positioned spatially in front of the target that are not yet masked.
    \item \textbf{Sequential Filtering:} Each candidate is rigorously tested against a set of exclusion rules. The agent is explicitly instructed to rule out environmental noise (e.g., dust, snow), surface artifacts (e.g., shadows, reflections), and self-occlusions before considering a candidate valid.
    \item \textbf{Justified Verdict:} Only candidates that survive the filtering stage are designated as valid missed occluders, accompanied by a logical justification for their inclusion.
\end{enumerate}

We model this verification process as a residual detection function. Let $\hat{I}$ denote the processed input image, where the currently identified occluders are masked (whited out) to prevent redundancy, while the rest of the scene context remains preserved. The agent performs inference on this view conditioned on the user query $q$:
\begin{equation}
    \Delta \mathcal{O} = \mathcal{A}_{ver}(\hat{I}, q \mid \mathcal{P}_{cot})
\end{equation}
If the agent identifies residual occluders ($\Delta \mathcal{O}$), the Segmentation Agent is recalled to ground these specific regions, updating the final mask set via Eq. \ref{eq:mask_union}.

\subsection{Diverse Hypothesis Generation}
\label{sec:ambiguity}

With the geometric constraints verified and refined, we now address the semantic dimension. A fundamental challenge in amodal completion is the inherent ambiguity of the invisible regions. For instance, a cat partially hidden by a sofa could plausibly be sleeping, sitting upright, or stretching. Conventional deterministic approaches typically commit to a single, arbitrary outcome, limiting the diversity of the results.

To address this, we leverage the extensive world knowledge embedded in MLLMs to explicitly model the latent semantic variable $S$ defined in Sec. \ref{sec:Problem_Formulation}. We employ the \textbf{Description Agent} ($\mathcal{A}_{desc}$), configured as a \textbf{Hypothesis Generator}, to reason about the scene context. To approximate the multimodal distribution of $S$, the agent is prompted to propose a set of $K$ diverse, plausible scenarios for the occluded content:
\begin{equation}
    \mathcal{H} = \{(T_k, w_k)\}_{k=1}^K = \mathcal{A}_{desc}(I_{obs}, q, K)
\end{equation}
where $T_k$ represents a distinct semantic description (e.g., specific pose or attribute), and $w_k$ is an estimated confidence score assigned by the agent, satisfying $\sum w_k = 1$. 

Crucially, each generated description $T_k$ serves as a holistic representation of the target object (e.g., ``a complete orange tabby cat with a long striped tail extending naturally''). As enforced by our prompt design, $T_k$ explicitly excludes any reference to the occluding objects (e.g., omitting ``behind the chair'') to prevent the Inpainting Agent from erroneously inferring occluder textures into the target's geometry. This structured output allows the framework to offer multiple interpretations of the same input. For automated execution, the hypothesis with the highest confidence ($T^*$) is selected.

\subsection{Final Plan Integration}
\label{sec:plan_integration}

Following the conclusion of the geometric verification (Sec. \ref{sec:refinement}) and the semantic inference (Sec. \ref{sec:ambiguity}), the framework proceeds to consolidate these distinct reasoning streams into a definitive execution plan, denoted as $\Pi$. 

The primary objective of this stage is to construct the conditioning inputs for the Inpainting Agent. To mitigate the interference of background clutter and strictly focus the generative prior on the object's geometry, we prepare a clean masked composite, $I_{input}$. This is achieved by isolating the visible pixels of the target (defined by the refined $M_{vis}$) while maintaining their original spatial layout within the potentially expanded image frame. Simultaneously, we explicitly suppress all extraneous regions, including both the original background scenes and the identified occluders, by replacing them with a neutral background color (e.g., white).

Consequently, the finalized plan $\Pi$, which is passed to the Inpainting Agent, consolidates three precise components: the pre-processed visual context $I_{input}$, the verified inpainting mask $M_{inpaint}$, and the selected semantic description $T^*$:
\begin{equation}
    \Pi = \{ I_{input}, M_{inpaint}, T^* \}
\end{equation}

\subsection{Visual Synthesis}
\label{sec:synthesis}

The final stage executes the reasoned plan $\Pi$ to generate the completed object. We employ an advanced inpainting model $\mathcal{G}$ as the Inpainting Agent. The generation is conditioned on the pre-processed visual context and the derived semantic guidance:
\begin{equation}
    I_{comp} = \mathcal{G}(z \mid I_{input}, M_{inpaint}, T^*)
\end{equation}
where $z$ denotes the initial latent noise. In this formulation, $M_{inpaint}$ serves as the binary mask defining the region to be filled, while $I_{input}$ provides the strictly preserved visible content. Since the complex geometric and semantic decisions are pre-determined in the planning phase, the inpainting model is not required to infer structure during synthesis. This allows it to produce high-fidelity results in a single stable pass, avoiding the error accumulation inherent in iterative pipelines.
\section{MAC-SCORE: A Human-Aligned Metric for Amodal Completion}

Evaluating amodal completion is challenging due to the lack of ground truth for occluded regions. The following discussion outlines the limitations of traditional metrics that motivate the design of our proposed MAC-Score.

\begin{figure}[t]
    \centering
    \includegraphics[width=1.0\linewidth]{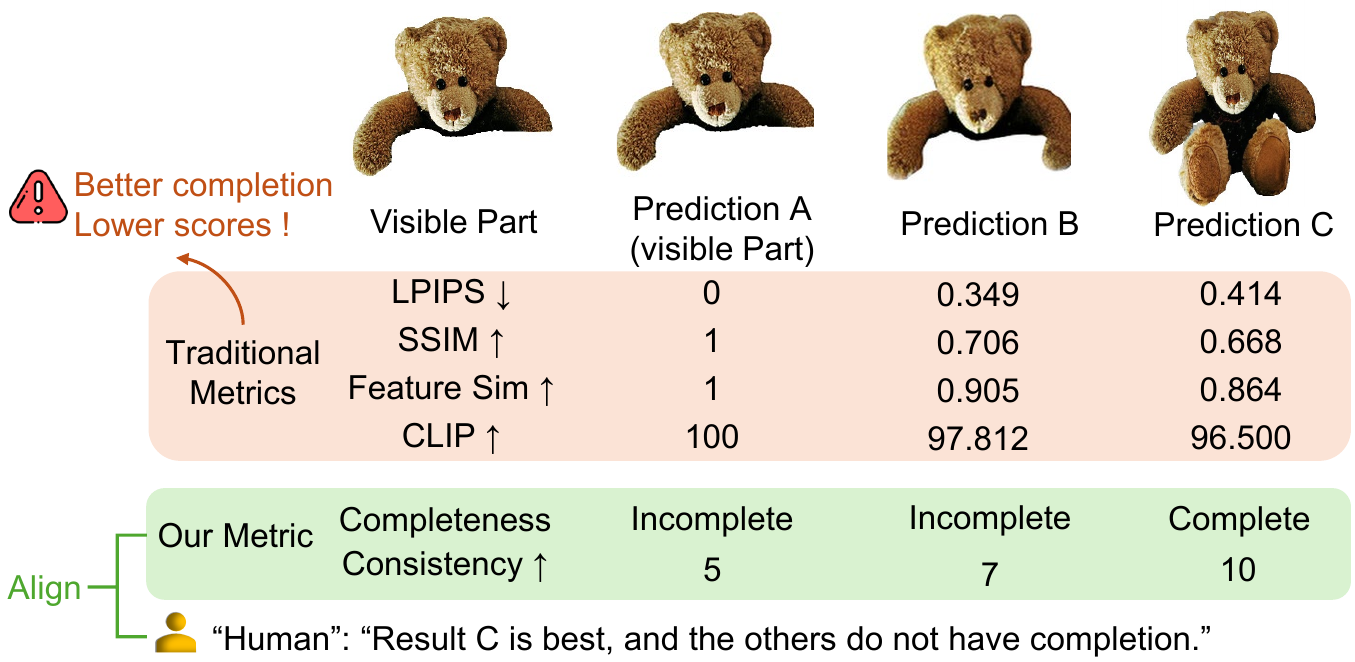}
    \caption{Comparison between traditional metrics and our MAC-Score. Traditional metrics fail to reflect human perception by assigning perfect scores to the incomplete \textbf{Prediction A} while penalizing the plausible \textbf{Prediction C} due to pixel deviations. In contrast, our \textbf{MAC-Score} correctly identifies Prediction C as the superior result with high consistency, aligning with human judgment.}
    \label{fig:metric_sample}
\end{figure}

\subsection{Limitations of Existing Metrics}
\label{sec:limitations}

Current quantitative metrics utilize a methodology that is fundamentally ill-suited for amodal completion. They typically calculate the distance or similarity between the original input image, which contains only visible parts, and the final completed result, which contains both visible and newly generated parts.
This reference-based mechanism leads to severe misalignment with human perception, as illustrated in Fig.~\ref{fig:metric_sample}.

\textbf{Reward for Incompleteness.} As shown in the comparison, Prediction A, which simply duplicates the original visible part without any completion, achieves a perfect LPIPS score of 0 and an SSIM of 1. This occurs because the metric treats the visible input as the ground truth, rewarding the absence of modification rather than the presence of completion.

\textbf{Penalty for Plausible Content.} The case of Prediction C reveals a critical failure. Prediction C represents a plausible and structurally complete teddy bear. However, traditional metrics assign it the worst scores (LPIPS 0.414, SSIM 0.668). The generated part, although semantically correct, are treated as errors because they deviate from the visible-only reference. Even Prediction B, which is partially complete but still fragmented, receives better scores than the fully complete result.

Consequently, existing metrics fail to capture the core objectives of the task: semantic consistency, structural completeness, and overall plausibility. 

\subsection{MAC-Score}

To address these limitations, we leverage MLLMs to establish a human-aligned evaluation framework. We configure the model as an expert \textit{AI evaluator} via structured prompting to assess completion quality through two complementary metrics. The model receives: (1) the original image, (2) the final completed object, and (3) the target object name.

\subsubsection{MAC-Completeness}

This metric targets the fundamental aspect of the task: whether the object is structurally complete.

\textbf{Prompting Strategy.} The MLLM is instructed to act as a visual perception expert. We define strict criteria where \textit{Complete} signifies that the object's natural and full structure is present, while \textit{Incomplete} indicates the object is truncated, missing parts, or distorted. The MLLM compares the completed result with the original image and outputs a structured JSON decision indicating whether the object is complete and the explanation. This process yields a clear, binary measure of whether the core goal of completion has been achieved.

\subsubsection{MAC-Consistency}

This metric evaluates the intrinsic structural and semantic coherence of the completed object relative to its visible parts, explicitly decoupling object quality from low-level rendering artifacts.

\textbf{Prompting Strategy.} Unlike traditional metrics which penalize any pixel-level deviation including lighting or slight misalignment, we instruct the MLLM to focus on the object's identity and structure. The prompt explicitly directs the evaluator to ignore low-level factors such as background blending, lighting shifts, or absolute position. Instead, it scores the completion from 0 to 10 based on three high-level dimensions:
\begin{enumerate}
    \item \textbf{Structural Continuity (0--4 points):} Assessing whether contours flow seamlessly and align naturally between the visible and generated regions, requiring geometric reasoning beyond pixel matching.
    \item \textbf{Semantic Consistency (0--4 points):} Verifies the generated parts are semantically correct for the specific object identity (e.g., a cat is not completed with a dog's tail).
    \item \textbf{Object Realism (0--2 points):} Judging whether the completed object, viewed in isolation, adheres to real-world physical plausibility.
\end{enumerate}

\textbf{Differentiation from Traditional Metrics.} This metric addresses a critical gap: traditional metrics often penalize valid completions due to pixel deviations.
By instructing the MLLM to disregard non-semantic factors, this metric provides a robust assessment of structural completeness and semantic consistency, closer to how humans perceive object permanence.
\section{Experiments}
\begin{figure*}[t]
    \centering
    \includegraphics[width=1.0\linewidth]{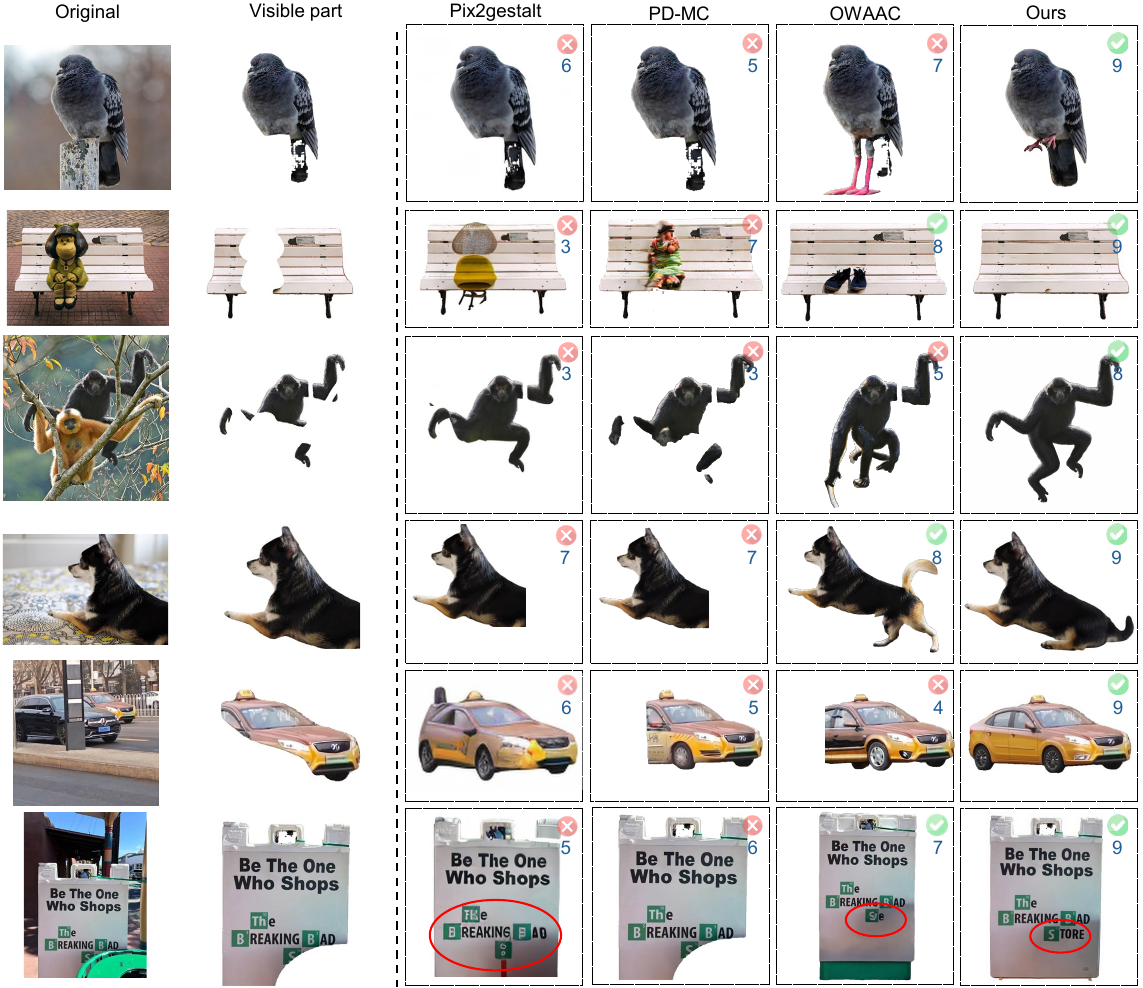}
    \caption{\textbf{Qualitative comparison against SOTA amodal completion approaches.} We compare with Pix2Gestalt~\cite{ozguroglu2024pix2gestalt}, PD-MC~\cite{xu2024amodal}, and OWAAC~\cite{ao2025open}. Annotations denote \textbf{MAC-Consistency} scores and \textbf{MAC-Completeness} states (\textcolor{green}{\checkmark}/\textcolor{red}{×}). \textbf{Row 1 (Anatomy):} Baselines truncate the tail or hallucinate unrealistic limbs (e.g., OWAAC's legs), while we recover the natural bird shape. \textbf{Row 2 (Texture):} Baselines leave significant ``ghost'' artifacts (yellow blobs), whereas we cleanly recover the texture. \textbf{Rows 3 \& 4 (Reasoning):} Baselines struggle with structure. In Row 3, PD-MC hallucinates an unnatural object; in Row 4, OWAAC misinterprets the dog's pose. Our method correctly infers both geometry and posture. \textbf{Row 5 (Geometry):} Competitors fail to extrapolate the taxi's length, resulting in distorted/truncated bodies; ours maintains the taxi's structural completeness. \textbf{Row 6 (Text):} Only our method accurately recovers missing semantic text characters. Overall, our superior visual quality aligns with the higher quantitative scores, verifying the rationality of our proposed metrics.}
    \label{fig:qualitative_comparison}
\end{figure*}

\subsection{Experimental Setup}

\subsubsection{Datasets}
Due to the absence of ground truth in real-world scenarios, our evaluation spans two domains: (i) open-world real-image benchmarks and (ii) a synthetic dataset with exact amodal ground truth for objective validation.

\textbf{Open-world benchmark.}
We follow the evaluation protocol of Ao et al.~\cite{ao2025open} on their dataset, which contains 2,379 images collected from four sources: Visual Genome (VG)~\cite{krishna2017visual} (1,234), a filtered COCO-A~\cite{zhu2017semantic} subset (751), copyright-free images from publicly accessible websites (228), and a LAION~\cite{schuhmann2021400m} subset (166).
Specifically, we use the provided target class label as the text query $q$ for each sample.

\textbf{HiFi-Amodal (ours).}
To better reflect contemporary daily-life imagery and challenging real-world conditions, we curate the HiFi-Amodal dataset, comprising approximately 220 images sourced from (i) self-captured photography and (ii) copyright-free public platforms.
Unlike existing benchmarks limited by low visual fidelity or uncurated composition, HiFi-Amodal emulates the high-quality, well-composed aesthetics of modern digital photography (e.g., social media aesthetics). Crucially, it introduces more challenging scenarios, including multiple instances of the same category, dense clutter, and diverse occlusion patterns.
Each image is annotated with a target query phrase $q$ for language-conditioned completion.
We will release the dataset to facilitate community research.

\textbf{Pix2Gestalt Occlusions~\cite{ozguroglu2024pix2gestalt}.}
For quantitative validation, we employ this synthetic benchmark containing paired ground-truth samples. We manually curate a refined subset of 300 instances, excluding low-fidelity cases (e.g., minimal target visibility) to ensure robust evaluation.

\subsubsection{Baselines}
We compare our proposed framework against several SOTA amodal completion methods, including:
\begin{itemize}
    \item \textbf{Pix2Gestalt}~\cite{ozguroglu2024pix2gestalt}: A diffusion-based approach that synthesizes plausible whole-object appearances from partially visible observations for amodal completion.
    \item \textbf{PD-MC}~\cite{xu2024amodal}: A progressive diffusion-based approach that iteratively completes occluded objects by leveraging mixed contextual cues.
    \item \textbf{OWAAC}~\cite{ao2025open}: A progressive open-world framework for amodal appearance completion of arbitrary objects.
\end{itemize}
All baselines are evaluated using their official codebases.

\subsubsection{Implementation Details}
Our framework is fully training-free and requires no task-specific fine-tuning.
\textbf{Reasoning Core.} We use OpenAI's GPT-4o~\cite{openai2024gpt4o_system_card} (via API) as the backbone for the Occlusion, Boundary, and Hypothesis agents. The Verification Agent uses Gemini 2.5 Pro~\cite{comanici2025gemini} (via API).
\textbf{Visual Modules.} We integrate X-SAM ~\cite{wang2025x} for prompt-based open-vocabulary segmentation and adopt FLUX-ControlNet-Inpainting~\cite{alimamacreative_flux_controlnet_inpainting} for the final high-fidelity appearance synthesis. All vision modules are run on NVIDIA A800 GPUs.

\subsubsection{Evaluation Metrics}
We evaluate all methods using both traditional metrics and our proposed evaluation metrics.

\textbf{Traditional Metrics.}
Following prior work~\cite{ao2025open}, since only the visible part of each target is available, we report standard similarity metrics as a proxy for quality by comparing the visible part of the target in the input image with the completed result. These include LPIPS~\cite{zhang2018unreasonable}, SSIM~\cite{wang2004image}, CLIP feature similarity~\cite{radford2021learning}, and VGG-16 perceptual feature similarity~\cite{gatys2016image}.

\textbf{MAC-Score.} We employ the open-source Qwen3-VL-32B-Instruct~\cite{bai2025qwen3vltechnicalreport} as an automated judge to assess holistic completion quality. This MLLM-based metric comprises two complementary scores: MAC-Completeness and MAC-Consistency.

\textbf{Human Evaluation Protocol.}
To validate perceptual quality, we conducted a user study with 50 participants, each evaluating 30 randomly sampled test cases from the HiFi-Amodal dataset. For each case, participants were presented with the input image, the query phrase $q$, and the randomized completion results from four methods. Participants evaluated three aspects: \textbf{Completeness}, a binary judgment (Yes/No) on whether the structure was fully recovered; \textbf{Consistency Score}, a 10-point scale assessing semantic and structural consistency with visible parts; and \textbf{Preference}, selecting the best result. In total, we collected 1{,}500 trials, yielding 6{,}000 completeness labels, 6{,}000 consistency ratings, and 1{,}500 preference votes. All participants provided informed consent.

\begin{table*}[t]
\caption{Quantitative comparison with SOTA methods on \textbf{Standard Benchmarks} (VG~\cite{krishna2017visual}, COCO-A~\cite{zhu2017semantic}, Free Images~\cite{ao2025open}, LAION~\cite{schuhmann2021400m}) and our \textbf{HiFi-Amodal Dataset}. We compare standard visible part consistency metrics (left) and our MAC-Score metrics (right). $\uparrow$: higher is better, $\downarrow$: lower is better. \textbf{Bold} indicates the best result.}
\label{tab:results_standard}
  \centering
    \begin{tabular}{llccccccc}
      \toprule
      \multirow{3}{*}{Dataset} & \multirow{3}{*}{Method} & \multicolumn{5}{c}{Standard Metrics} & \multicolumn{2}{c}{MAC-Score} \\
      \cmidrule(lr){3-7} \cmidrule(lr){8-9}
      & & CLIP (Img) & CLIP (Txt) & Visual & Semantic & Structural & MAC & MAC \\
      & & Score & Score & Consistency & Consistency & Consistency & Completeness &Consistency  \\
      \cmidrule(lr){3-3} \cmidrule(lr){4-4} \cmidrule(lr){5-5} \cmidrule(lr){6-6} \cmidrule(lr){7-7} \cmidrule(lr){8-8} \cmidrule(lr){9-9}
      & & $\uparrow$ & $\uparrow$ & $\downarrow$ LPIPS & $\uparrow$ Feature Sim. & $\uparrow$ SSIM & $\uparrow$ (\%) & $\uparrow$ (1-10) \\
      \midrule

      \multirow{4}{*}{VG~\cite{krishna2017visual}}
        & PD-MC\cite{xu2024amodal}       & \textbf{94.360} & 28.367 & 0.578 & 0.413 & 0.463 &41.111 &6.811 \\
        & Pix2gestalt\cite{ozguroglu2024pix2gestalt} & 88.985& 27.672 & 0.429 & 0.554 & 0.726 & 34.234& 6.642\\
        & OWAAC\cite{ao2025open}         &91.988 & \textbf{28.470} & 0.310 & 0.658 & 0.732 & 63.214&7.514 \\
        \rowcolor{highlightgray}
        & Ours                           & 90.935& 28.377 & \textbf{0.217} & \textbf{0.859} & \textbf{0.836} & \textbf{67.956}& \textbf{7.861}\\
      \midrule

      \multirow{4}{*}{COCO-A~\cite{zhu2017semantic}}
        & PD-MC\cite{xu2024amodal}       & \textbf{94.167}& 27.383 & 0.664 & 0.328 & 0.382 & 34.319& 6.364\\
        & Pix2gestalt\cite{ozguroglu2024pix2gestalt} & 88.801& 26.998 & 0.471 & 0.524 & 0.695 & 30.414 & 6.351 \\
        & OWAAC\cite{ao2025open}         &91.171 & \textbf{27.612} & 0.351 & 0.609 & 0.718 & 52.899 & 6.969 \\
        \rowcolor{highlightgray}
        & Ours                           & 89.134& 27.516 & \textbf{0.286} & \textbf{0.822} & \textbf{0.804} & \textbf{58.934} & \textbf{7.576} \\
      \midrule

      \multirow{4}{*}{Free Images~\cite{ao2025open}}
        & PD-MC\cite{xu2024amodal}       & \textbf{96.005}& 28.333 & 0.720 & 0.279 & 0.309 & 25.301 & 6.333 \\
        & Pix2gestalt\cite{ozguroglu2024pix2gestalt} & 89.464& 27.621 & 0.393 & 0.613 & 0.732 & 25.301 & 6.385\\
        & OWAAC\cite{ao2025open}         &91.928 & \textbf{28.652} & 0.269 & 0.698 & 0.753 & 59.036 & 7.610\\
        \rowcolor{highlightgray}
        & Ours                           &90.378 & 28.569 & \textbf{0.235} & \textbf{0.836} & \textbf{0.828} & \textbf{69.477} & \textbf{8.092} \\
      \midrule

      \multirow{4}{*}{LAION~\cite{schuhmann2021400m}}
        & PD-MC\cite{xu2024amodal}       &\textbf{94.687} & 27.573 & 0.692 & 0.299 & 0.346 & 33.898 & 6.468\\
        & Pix2gestalt\cite{ozguroglu2024pix2gestalt} & 88.696& 27.260 & 0.467 & 0.527 & 0.691 & 41.242 & 6.903\\
        & OWAAC\cite{ao2025open}         &90.795 & \textbf{28.123} & \textbf{0.319} & 0.657 & \textbf{0.751} & 63.841 & 7.734\\
        \rowcolor{highlightgray}
        & Ours                           &89.939 & 27.966& 0.337 & \textbf{0.837} & 0.750 & \textbf{74.011} & \textbf{8.118}\\
      \bottomrule
    \end{tabular}
\end{table*}

\begin{table*}[t]
\caption{\textbf{Comprehensive Evaluation on HiFi-Amodal Dataset.} We integrate results from standard automated metrics, our MAC-Score, and the large-scale User Study (Human Eval).
For PD-MC~\cite{xu2024amodal}, we report both ``Valid Only'' (successful completions) and ``Full Set'' (failures substituted with input).
$\uparrow$: higher is better, $\downarrow$: lower is better. \textbf{Bold} indicates the best result.}
\label{tab:hifi_user_combined}
\centering
\footnotesize
\begin{tabular*}{\linewidth}{@{}l@{\extracolsep{\fill}}cccccccccc@{}}
\toprule
\multirow{3}{*}{Method} & \multicolumn{5}{c}{Standard Metrics} & \multicolumn{2}{c}{MAC-Score} & \multicolumn{3}{c}{Human Evaluation (User Study)} \\
\cmidrule(lr){2-6} \cmidrule(lr){7-8} \cmidrule(lr){9-11}
 & \textit{CLIP-I} $\uparrow$ & \textit{CLIP-T} $\uparrow$ & \textit{LPIPS} $\downarrow$ & \textit{Feat.} $\uparrow$ & \textit{SSIM} $\uparrow$ & \textit{MAC-Comp.} $\uparrow$ & \textit{MAC-Cons.} $\uparrow$ & \textit{Comp.} $\uparrow$ & \textit{Cons.} $\uparrow$ & \textit{Pref.} $\uparrow$ \\
\midrule
Pix2gestalt\cite{ozguroglu2024pix2gestalt} & 91.996 & 27.101 & 0.192 & 0.837 & 0.876 & 38.144 & 6.794 & 45.683 & 5.312 & 15.324 \\
\multicolumn{11}{@{}l}{\textit{PD-MC\cite{xu2024amodal}:}} \\
\hspace{3mm} Valid Only & 85.822 & 26.467 & 0.450 & 0.763 & 0.719 & 16.857 & 5.490 & - & - & - \\
\hspace{3mm} Full Set   & \textbf{95.290} & 27.019 & \textbf{0.153} & \textbf{0.921} & \textbf{0.904} & 21.254 & 5.210 & 15.732 & 3.879 & 5.175 \\ 
OWAAC\cite{ao2025open}                     & 87.014 & 27.320 & 0.409 & 0.787 & 0.754 & 30.882 & 6.573 & 18.152 & 4.335 & 7.373 \\
\textbf{Ours}                              & 90.979 & \textbf{27.553} & 0.235 & 0.842 & 0.844 & \textbf{65.454} & \textbf{8.023} & \textbf{74.113} & \textbf{6.861} & \textbf{72.128} \\
\bottomrule
\end{tabular*}
\end{table*}

\subsection{Main Results and Comparisons}

\subsubsection{Qualitative Comparison}
Figure~\ref{fig:qualitative_comparison} demonstrate that our method produces more complete and semantically consistent amodal completions across diverse occlusion and truncation cases. Compared with Pix2Gestalt~\cite{ozguroglu2024pix2gestalt}, PD-MC~\cite{xu2024amodal}, and OWAAC~\cite{ao2025open}, our framework reduces common failure modes such as occluder ghosting/color bleeding, fragmented completions, pose drift, and text scrambling.
Specifically, Pix2Gestalt frequently suffers from residual occluder artifacts and texture/color leakage (e.g., the yellow/gray ghost regions on the bench in Row~2), and its text completion can become scrambled or illegible (Row~6). 
Progressive baselines are more prone to semantic drift or inference instability under challenging occlusions. In particular, PD-MC may terminate prematurely, producing conservative outputs that only slightly extend the visible parts (e.g., Row~3). OWAAC can change the object’s pose and violate the visible-part geometry (e.g., completing a lying dog into a standing pose in Row~4). In contrast, our method preserves the visible structure while synthesizing plausible missing regions, yielding coherent object shape completion (Rows~1--5) and more legible semantic-detail recovery (e.g., restoring the word ``STORE'' in Row~6).

\subsubsection{Quantitative Comparison on Traditional and MAC-Score Metrics}

We evaluate amodal completion with two sets of metrics (Tables~\ref{tab:results_standard} and~\ref{tab:hifi_user_combined}). Traditional visible-part similarity metrics use the input visible part as the reference and quantify how well the output preserves the observed region. While these scores reflect visible-region fidelity, they are not well suited to amodal completion: they do not assess whether the inferred (previously invisible) parts are structurally complete and semantically consistent, and they can sometimes favor conservative outputs that make minimal changes. This limitation is clearly exemplified by the PD-MC results in Table~\ref{tab:hifi_user_combined}. Substituting failure cases with the original input artificially boosts visible-part metrics, notably improving LPIPS from 0.450 to 0.153, despite the absence of valid completion. In contrast, both MAC-Completeness (21.254\%) and human completeness ratings (15.732\%) remain low, accurately reflecting the poor performance masked by traditional scores.

Focusing on these human-aligned metrics, our method demonstrates superior amodal recovery. It achieves the highest MAC-Score across all open-world benchmarks (Table~\ref{tab:results_standard}) and the HiFi-Amodal dataset (Table~\ref{tab:hifi_user_combined}), attaining 65.454\% Completeness and 8.023 Consistency. These results align with the user study, providing quantitative evidence that our reasoning-driven framework significantly improves amodal completeness and semantic consistency where baselines fall short.

\subsubsection{User Study Analysis}
\label{sec:user_study}

We conducted a user study with 50 participants to evaluate perceptual quality (Table~\ref{tab:hifi_user_combined}).

\textbf{Human Preference.} Our framework achieves a dominant User Preference Ratio of 72.128\%, surpassing the second-best method, Pix2Gestalt~\cite{ozguroglu2024pix2gestalt} (15.324\%), by a substantial margin. This significant lead demonstrates that human evaluators perceive our reasoning-driven completions as markedly more plausible in complex, open-vocabulary scenarios.

\textbf{Completeness and Consistency.}
Our method achieves a Completeness Rate of 74.113\%, significantly outperforming the nearest baseline, Pix2Gestalt~\cite{ozguroglu2024pix2gestalt} (45.683\%). This highlights our advantage in inferring invisible structures, whereas baselines often conservatively truncate objects. Additionally, we lead in Visual Consistency with a score of 6.86 (vs. 5.31 for Pix2Gestalt and $<4.5$ for others), confirming that our framework generates textures and geometries that are both structurally complete and seamlessly coherent.

\subsubsection{Objective Comparison on Ground-Truth Benchmark}

\begin{table*}[t]
\caption{Comprehensive quantitative comparison on the Ground-Truth Benchmark. \textbf{Note on PD-MC~\cite{xu2024amodal}:} ``Valid Only'' denotes the subset of successfully completed cases. For the ``Full Set'', failure cases are substituted with the original visible inputs. This standardization yields perfect scores for these samples, artificially inflating visible consistency metrics (e.g., Vis-SSIM~\cite{wang2004image}) despite the lack of effective completion. $\uparrow$: higher is better, $\downarrow$: lower is better. \textbf{Bold} indicates the best performance.}
\label{tab:gt_comprehensive}
\centering
\resizebox{\linewidth}{!}{%
\begin{tabular}{lccccccccccc}
\toprule
\multirow{2}{*}{Method} & \multicolumn{5}{c}{Evaluation against Ground Truth (Invisible)} & \multicolumn{3}{c}{Evaluation against Visible Input (Visible)} & \multicolumn{2}{c}{MAC-Score} \\
\cmidrule(lr){2-6} \cmidrule(lr){7-9} \cmidrule(lr){10-11}
 & \textit{GT-LPIPS} $\downarrow$ & \textit{GT-Feat.} $\uparrow$ & \textit{GT-SSIM} $\uparrow$ & \textit{GT-PSNR} $\uparrow$ & \textit{GT-mIoU} $\uparrow$ & \textit{Vis-LPIPS} $\downarrow$ & \textit{Vis-Feat.} $\uparrow$ & \textit{Vis-SSIM} $\uparrow$ & \textit{MAC-Comp.} $\uparrow$ & \textit{MAC-Cons.} $\uparrow$ \\
\midrule
Pix2Gestalt \cite{ozguroglu2024pix2gestalt} & 0.197 & 0.812 & 0.843 & 17.453 & 0.712 & 0.095 & 0.906 & 0.910 & 44.326 & 7.279 \\
\multicolumn{11}{l}{\textit{PD-MC\cite{xu2024amodal}:}} \\
\hspace{3mm} Valid Only$^\dagger$ & 0.279 & 0.751 & 0.817 & 16.220 & 0.537 & 0.216 & 0.820 & 0.857 & 16.578 & 5.333 \\
\hspace{3mm} Full Set             & 0.225 & 0.781 & 0.859 & 16.518 & 0.608 & \textbf{0.060} & \textbf{0.950} & \textbf{0.960} & 21.595 & 5.732 \\
OWAAC \cite{ao2025open}                     & 0.311 & 0.784 & 0.757 & 13.642 & 0.436 & 0.194 & 0.800 & 0.871 & 25.658 & 5.833 \\
\midrule
\textbf{Ours}                               & \textbf{0.150} & \textbf{0.858} & \textbf{0.885} & \textbf{19.034} & \textbf{0.748} & 0.113 & 0.894 & 0.934 & \textbf{63.646} & \textbf{7.489} \\
\bottomrule
\end{tabular}%
}
\end{table*}

We further validate our method by comparing generated objects against ground truth on the Pix2Gestalt dataset~\cite{ozguroglu2024pix2gestalt}. Our framework achieves SOTA performance on all invisible-region metrics. Specifically, we achieve the lowest perceptual error (GT-LPIPS = 0.150) and the highest structural similarity (GT-SSIM = 0.885).
A notable anomaly occurs with PD-MC~\cite{xu2024amodal}: substituting its failure cases with original inputs artificially inflates visible metrics (e.g., Vis-LPIPS 0.060). This exemplifies the phenomenon where traditional metrics reward conservative incompleteness. Despite these misleading scores, PD-MC performs poorly on actual invisible regions (GT-LPIPS 0.225; MAC-Completeness 21.59\%), whereas our method delivers robust completion across all dimensions.

\subsection{Validation of the Proposed Evaluation Protocol}
\label{sec:validation_metrics}

\subsubsection{Correlation with Human Preferences}

To validate our MLLM-based paradigm, we calculate Spearman Rank Correlations with user study data. As shown in Table~\ref{tab:human_correlation}, traditional metrics (e.g., LPIPS, CLIP) exhibit weak or negative correlations, confirming that pixel fidelity fails to capture object completeness. In contrast, our metrics demonstrate robust alignment: MAC-Completeness achieves the highest correlation ($\rho = 0.516$), serving as a strong predictor of completeness, while MAC-Consistency also aligns significantly with human perception ($\rho = 0.490$). This confirms that our paradigm effectively mirrors human perceptual preferences.

\begin{table}[t]
\caption{Correlation between automated metrics and human judgments (Spearman's $\rho$). Our metrics demonstrate robust alignment with human perception.}
\label{tab:human_correlation}
\centering
\footnotesize 
\begin{tabular*}{\linewidth}{@{\extracolsep{\fill}}lcc}
\toprule
\multirow{2}{*}{Metric (Input)} & \multicolumn{2}{c}{Correlation with Human Judgments ($\rho$)} \\
\cmidrule(lr){2-3}
 & Human Completeness & Human Consistency \\
\midrule
LPIPS (Visible)    & 0.297  & 0.170 \\
SSIM (Visible)     & -0.310 & -0.194 \\
VGG Sim. (Visible) & -0.380 & -0.158 \\
CLIP Text Score         & -0.322 & -0.098 \\
\midrule
\textbf{MAC-Completeness} & \textbf{0.516} & 0.433 \\
\textbf{MAC-Consistency}  & 0.473 & \textbf{0.490} \\
\bottomrule
\end{tabular*}
\end{table}

\subsubsection{Correlation with Ground-Truth Metrics}

We analyzed correlations with ground-truth on the Pix2Gestalt dataset~\cite{ozguroglu2024pix2gestalt}. As shown in Table~\ref{tab:gt_correlation}, traditional metrics reveal severe limitations. While Vis-SSIM correlates moderately with GT-SSIM ($\rho=0.496$), it fails to predict perceptual and semantic quality, showing negligible or negative correlations with GT-LPIPS, mIoU, and Feature Similarity. This confirms the limitations discussed in Sec.~\ref{sec:limitations}: pixel fidelity in visible regions does not guarantee valid amodal completion.

In contrast, our MAC-Score demonstrate robust predictive power. Specifically, MAC-Completeness serves as the strongest predictor for semantic and perceptual quality, achieving the highest alignment with Feature Similarity ($r_{pb}=0.486$) and a strong negative correlation with GT-LPIPS($r_{pb}=-0.372$). Meanwhile, MAC-Consistency proves most effective for structural recovery, yielding the highest correlation with mIoU ($r_{pb}=0.484$). This validates our paradigm as an effective proxy for the core objectives of amodal completion.

\begin{table}[t]
\caption{Correlation with GT measures ($r_{pb}$/$\rho$). While Vis-SSIM aligns with structure, our MAC metrics best predict perceptual (LPIPS), shape (mIoU), and semantic (Feat.) fidelity.}
\label{tab:gt_correlation}
\centering
\footnotesize 
\setlength{\tabcolsep}{0pt} 

\begin{tabular*}{\linewidth}{@{\extracolsep{\fill}}lccccc}
\toprule
\multirow{2}{*}{Predictor} & \multicolumn{5}{c}{Correlation with GT Measures} \\
\cmidrule(lr){2-6}
 & \textit{LPIPS} $\downarrow$ & \textit{SSIM} $\uparrow$ & \textit{PSNR} $\uparrow$ & \textit{mIoU} $\uparrow$ & \textit{Feat.} $\uparrow$\\
\midrule
Vis-LPIPS    & 0.143 & -0.304 & -0.133 & -0.013 & 0.060 \\
Vis-SSIM     & -0.248 & \textbf{0.496} & 0.255 & -0.027 & 0.034 \\
Vis-Feat.    & -0.049 & 0.152 & 0.037 & 0.079 & -0.060 \\
\midrule
\textbf{MAC-Completeness} & \textbf{-0.372} & 0.183 & \textbf{0.315} & 0.455 & \textbf{0.486} \\
\textbf{MAC-Consistency}  & -0.240 & 0.040 & 0.186 & \textbf{0.484} & 0.366 \\
\bottomrule
\end{tabular*}
\end{table}

\begin{table}[t]
\caption{\textbf{Ablation studies.} We validate the contribution of individual agents, the reasoning backbone, and the visual synthesis agent within the proposed framework.}
\label{tab:ablation}
\centering
\footnotesize
\setlength{\tabcolsep}{1.5pt}
\begin{tabular*}{\linewidth}{@{\extracolsep{\fill}}lccccc}
\toprule
Variant & M-Com. $\uparrow$ & M-Con. $\uparrow$ & Feat. $\uparrow$ & SSIM $\uparrow$ & LPIPS $\downarrow$ \\
\midrule
\textit{Component Ablation} & & & & & \\
\hspace{3mm} w/o Description Agent   & 64.717 & 7.951 & 0.839 & 0.843 & 0.237 \\
\hspace{3mm} w/o Boundary Agent      & 45.547 & 7.755 & 0.850 & \textbf{0.865} & 0.224 \\ 
\hspace{3mm} w/o Verification Agent  & 57.727 & 7.892 & 0.855 & 0.849 & 0.230 \\
\hspace{3mm} w/ Deterministic Desc.  & \textbf{65.731} & 8.012 & 0.846 & 0.846 & 0.234 \\
\midrule
\textit{Backbone Replacement} & & & & & \\
\hspace{3mm} w/ SDXL Backbone        & 49.575 & 7.671 & 0.855 & 0.853 & 0.222 \\
\hspace{3mm} w/ Qwen3-VL (OpenSrc.)  & 60.000 & 7.989 & \textbf{0.862} & 0.858 & \textbf{0.211} \\ 
\midrule
\textbf{Ours (Full Framework)}       & 65.454 & \textbf{8.023} & 0.842 & 0.844 & 0.235 \\
\bottomrule
\end{tabular*}
\end{table}

\subsection{Ablation Studies}
\label{sec:ablation}

To validate the contribution of individual components within our Collaborative Multi-Agent Reasoning Framework, we conducted a comprehensive ablation study on the HiFi-Amodal dataset. Quantitative results are summarized in Table~\ref{tab:ablation}.

\subsubsection{Effectiveness of Collaborative Agents}
We analyze the impact of specific functional modules. Results underscore the critical role of our self-correcting mechanism: removing the Verification Agent drops MAC-Completeness from 65.45\% to 57.73\%. This confirms that closed-loop CoT refinement is indispensable for resolving residual occluders and ensuring geometric integrity. Similarly, removing the Boundary Analysis Agent causes a sharp decline to 45.55\% in completeness, validating the necessity of dynamic canvas expansion. 
Regarding semantic guidance, the full framework achieves comparable MAC-Consistency to the deterministic variant. While the deterministic approach favors generic completions, our diverse semantic reasoning is critical for modeling inherent ambiguity without compromising semantic consistency.

\subsubsection{Modularity and Model Generalization}
A key advantage of our framework lies in its decoupled design, allowing for the flexible replacement of backbones. To investigate generalization, we replaced the proprietary GPT-4o~\cite{openai2024gpt4o_system_card} with the open-source Qwen3-VL-32B~\cite{bai2025qwen3vltechnicalreport}. Despite a performance gap (Completeness: 60.00\% vs. 65.45\%), it significantly outperforms non-reasoning baselines, demonstrating that efficacy stems from our collaborative methodology rather than dependence on a closed-source backbone. Moreover, the performance gain observed when upgrading the visual backbone from SDXL~\cite{podell2023sdxl} to FLUX~\cite{alimamacreative_flux_controlnet_inpainting} highlights the framework's extensibility. This ``plug-and-play'' modularity ensures the framework is future-proof, capable of seamlessly integrating advanced foundation models without architectural overhaul.

\begin{figure*}[t]
    \centering
    \includegraphics[width=0.98\textwidth]{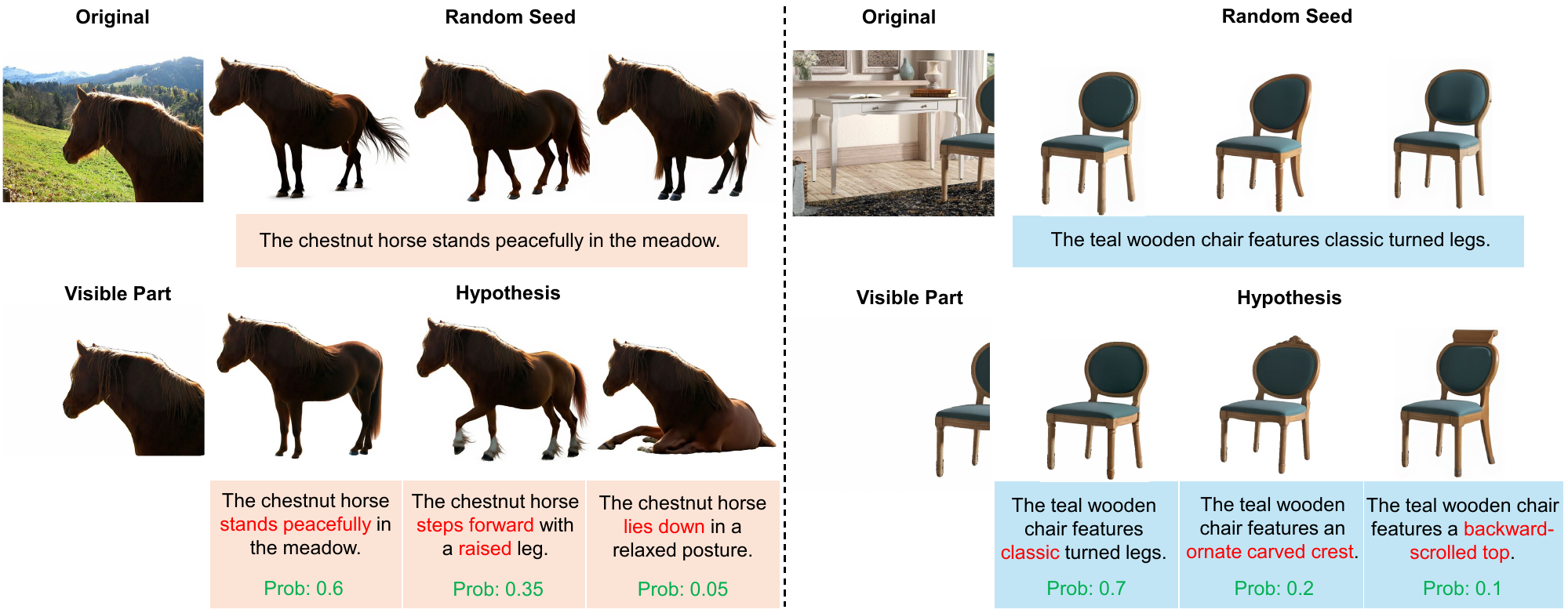} 
    \caption{\textbf{Visualizing Ambiguity-Aware Hypothesis Generation.} 
    \textbf{Top:} Standard stochastic sampling collapses to a single semantic mode, yielding only trivial texture variations. 
    \textbf{Bottom:} Our framework explicitly reasons about invisible semantics, generating diverse interpretations (e.g., \textit{Stepping} vs. \textit{Lying Down}) with estimated confidence ($Prob$). This transforms completion from static inpainting into controllable, probabilistic exploration.}
    \label{fig:ambiguity_vis}
\end{figure*}

\subsection{Visualizing Diverse Hypothesis Generation}
\label{sec:ambiguity_vis}

A fundamental premise of our work is that amodal completion is ill-posed: a single occluded input often admits multiple valid interpretations. While standard generative approaches rely on stochastic seed variation, this typically yields only low-level textural noise rather than meaningful semantic diversity. Figure~\ref{fig:ambiguity_vis} contrasts our approach against a stochastic baseline.

\subsubsection{Stochasticity vs. Semantics} As shown in the top row of Figure~\ref{fig:ambiguity_vis}, changing the random seed for Inpainting Agent results in outputs that converge to a single semantic mode, limited to minor pixel-level fluctuations. In contrast, our Hypothesis Generator (Bottom Row) explicitly models the semantic solution space. By prompting the MLLM to hypothesize diverse scenarios based on visual context, we generate structurally distinct outcomes. For instance, given the visible upper body of a horse, our agent reasons that it could plausibly be ``stepping forward,'' ``grazing,'' or ``standing still.''

\subsubsection{Quantitative Diversity Analysis}
We quantify this difference in Table~\ref{tab:diversity}. Over 5 variations on 100 samples, our diverse hypothesis generation yields clear gains over random-seed sampling, boosting visual diversity (pairwise LPIPS) by 19.6\% and semantic diversity (pairwise CLIP distance) by 46.3\%, demonstrating genuinely controllable and meaningfully distinct completions.

\begin{table}[t]
\caption{Quantitative analysis of generation diversity based on average pairwise distance between variations.}
\label{tab:diversity}
\centering
\resizebox{\linewidth}{!}{%
\small
\begin{tabular}{lcc}
\toprule
Method Strategy & Visual Diversity & Semantic Diversity \\
 & (Pairwise LPIPS $\uparrow$) & (Pairwise CLIP Dist. $\uparrow$) \\
\midrule
Baseline (Random Seeds) & 0.060 & 0.030 \\
\textbf{Ours} & \textbf{0.071} & \textbf{0.043} \\
\midrule
\textit{Relative Gain} & \textbf{\textit{+19.6\%}} & \textbf{\textit{+46.3\%}} \\
\bottomrule
\end{tabular}
}
\end{table}

\begin{table}[t]
\caption{Cross-model validation. Spearman's $\rho$ confirms that the open-source Qwen3-VL-32B achieves robust human alignment comparable to proprietary SOTA models.}
\label{tab:model_comparison}
\centering
\footnotesize
\begin{tabular}{lcc}
\toprule
Evaluator Backbone & Completeness ($\rho$) & Consistency ($\rho$) \\
\midrule
GPT-4o ~\cite{openai2024gpt4o_system_card}           & 0.493 & \textbf{0.534} \\
Gemini-2.5-Flash ~\cite{comanici2025gemini}  & \textbf{0.547} & 0.464 \\
Qwen3-VL-32B-Instruct~\cite{bai2025qwen3vltechnicalreport} & 0.516 & 0.490 \\
\bottomrule
\end{tabular}
\end{table}

\subsection{Cross-Backbone Validation and Evaluation Reproducibility}
\label{sec:opensource_generalization}

To ensure reproducibility, we adopt the open-source Qwen3-VL-32B-Instruct~\cite{bai2025qwen3vltechnicalreport} as our default evaluator, replicate the evaluation with SOTA MLLMs (GPT-4o~\cite{openai2024gpt4o_system_card}, Gemini-2.5-Flash~\cite{comanici2025gemini}) to validate reliability. As shown in Table~\ref{tab:model_comparison}, strong correlations across backbones confirm our metric is model-agnostic. Notably, Qwen3-VL-32B~\cite{bai2025qwen3vltechnicalreport} demonstrates competitive alignment with human judgment, even surpassing GPT-4o~\cite{openai2024gpt4o_system_card} in MAC-Completeness ($\rho=0.516$ vs. $0.493$). Stability tests (temperature=0) further verify reproducibility, yielding negligible fluctuations for MAC-Completeness ($\pm 0.7\%$ Std) and MAC-Consistency ($\pm 0.026$ Std).
      
\section{Conclusion}
In this work, we present a Reasoning-Driven Multi-Agent Framework that reframes amodal completion by decoupling semantic planning from visual synthesis, addressing the inference instability and error accumulation of progressive approaches. By integrating a CoT Verification Agent for closed-loop self-correction and a Hypothesis Generator for modeling semantic ambiguity, our framework achieves superior completion quality and meaningful diversity. We also introduce the MAC-Score, establishing a robust, human-aligned standard to resolve the long-standing mismatch between conventional perceptual metrics and the requirements of amodal completion. Extensive experiments demonstrate that our holistic paradigm significantly outperforms SOTA methods. 
However, the collaborative interaction involving multiple MLLM agents and verification steps incurs higher computational costs compared to baselines. Future work could address this via model distillation, condensing these reasoning capabilities into a lightweight network for efficient inference.



\bibliographystyle{IEEEtran}
\bibliography{IEEEabrv, main}


\begin{IEEEbiography}[{\includegraphics[width=1in,height=1.25in,clip,keepaspectratio]{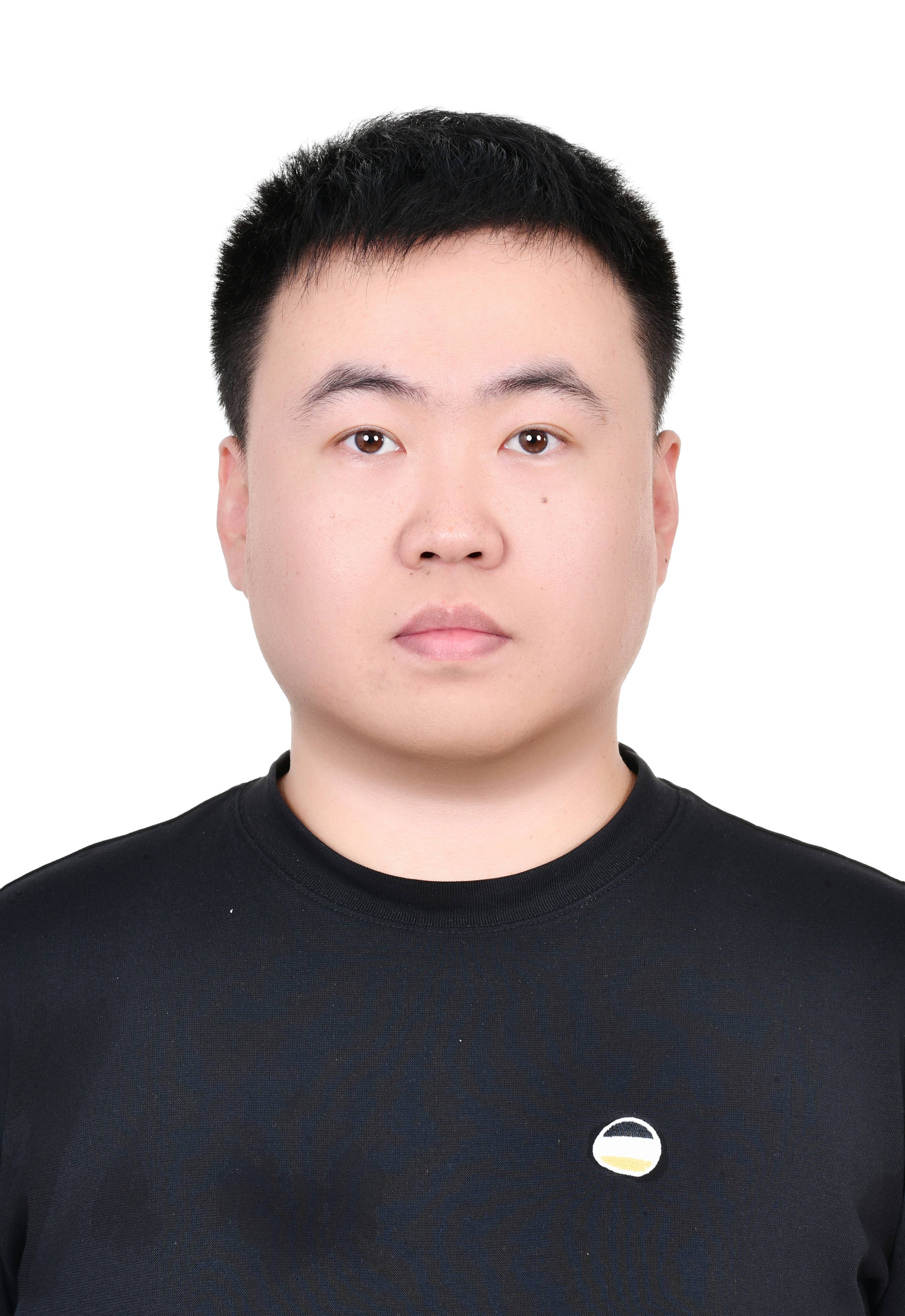}}]{Hongxing Fan}
received the B.Eng. degree from the North University of China in 2017, and the M.Eng. degree from the University of Chinese Academy of Sciences in 2021. He is currently pursuing the Ph.D. degree at Beihang University. His research interests include personalized image generation and 3D content creation.
\end{IEEEbiography}
\begin{IEEEbiography}[{\includegraphics[width=1in,height=1.25in,clip,keepaspectratio]{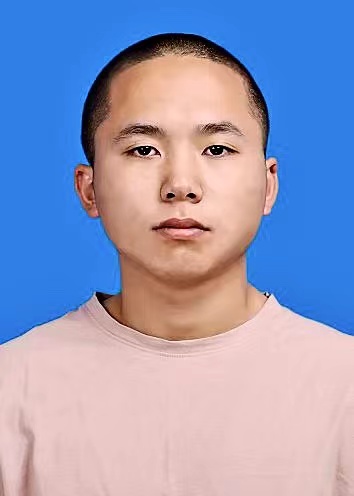}}]{Shuyu Zhao} is an undergraduate student at the School of Software, Beihang University. 
His research interests lie in artificial intelligence and 3D, particularly in amodal completion.
\end{IEEEbiography}
\begin{IEEEbiography}[{\includegraphics[width=1in,height=1.25in,clip,keepaspectratio]{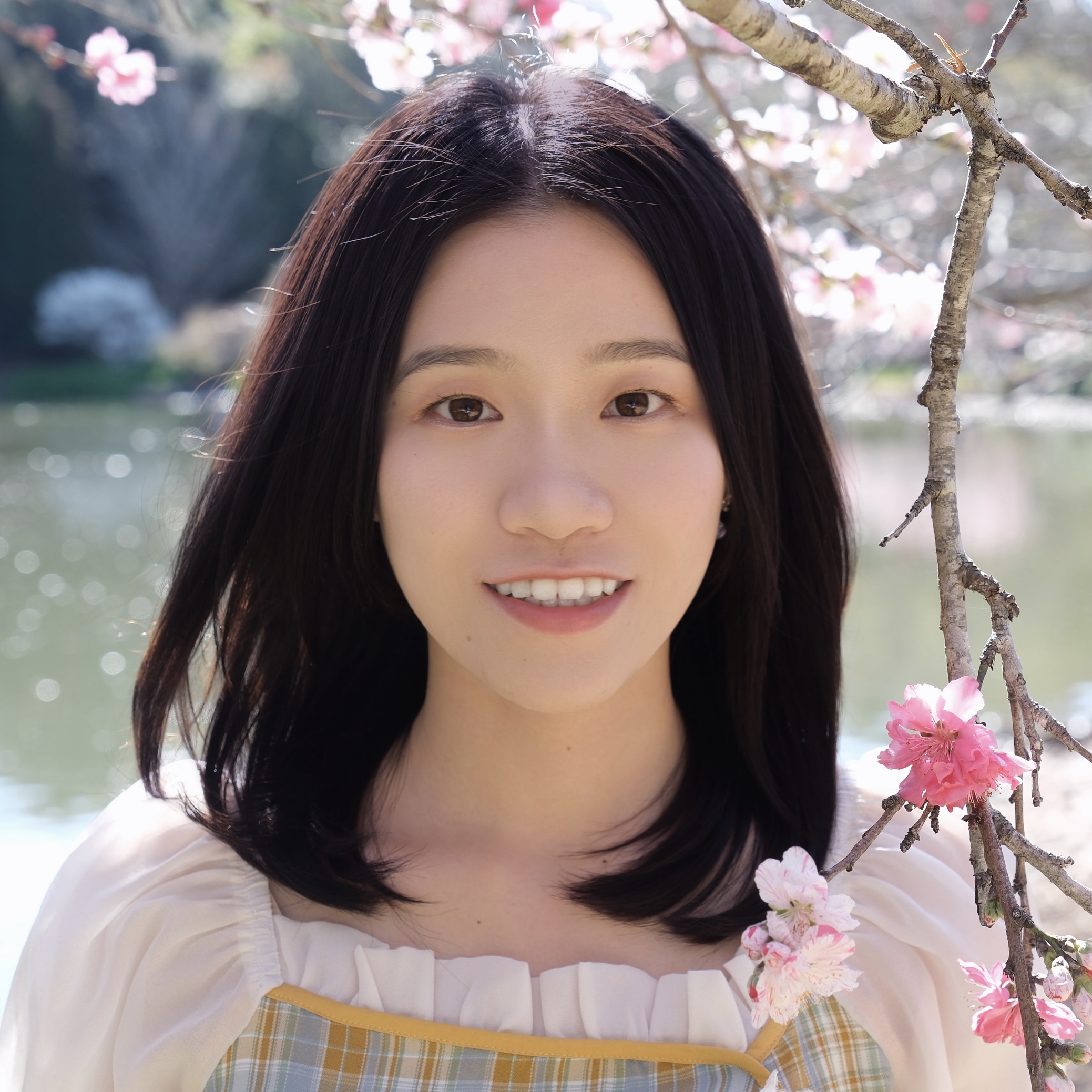}}]{Jiayang Ao} received the PhD degree from the University of Melbourne in 2025. Her work focuses on bridging the gap between computer vision systems and human perception, with expertise in understanding and completing visually occluded content. 
\end{IEEEbiography}
\begin{IEEEbiography}[{\includegraphics[width=1in,height=1.25in,clip,keepaspectratio]{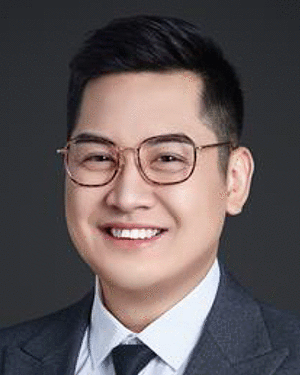}}]{Lu Sheng} (Member, IEEE) received the BE degree from Zhejiang University, China, in 2011, and the PhD degree from The Chinese University of Hong Kong, Hong Kong, in 2016. From 2017 to 2019, he was a post-doctoral researcher with the Multimedia Laboratory (MMLab), The Chinese University of Hong Kong. He is currently a Professor with the School of Software, Beihang University, China. His research interests include 3D computer vision and embodied AI, particularly focusing on developing generalizable models for understanding, interacting with and synthesizing the 3D/4D visual world.
\end{IEEEbiography}



\end{document}